\pdfoutput=1

\documentclass[11pt]{article}

\usepackage{emnlp2021}
\usepackage{graphicx}
\usepackage{times}
\usepackage{booktabs}
\usepackage{latexsym}
\usepackage{amsmath}
\usepackage{amssymb}
\usepackage{multirow}
\usepackage[T1]{fontenc}

\usepackage[utf8]{inputenc}

\usepackage{microtype}
\usepackage{color}

\title{HAN: Higher-order Attention Network for \\ Spoken Language Understanding}

\author{
Dongsheng Chen, Zhiqi Huang, Yuexian Zou\\
	Peking University, China \\
	chends@stu.pku.edu.cn,
\{zhiqihuang, zouyx\}@pku.edu.cn}

\begin{document}
\maketitle~

\begin{abstract}
Spoken Language Understanding (SLU), including intent detection and slot filling, is a core component in human-computer interaction.
The natural attributes of the relationship among the two subtasks make higher requirements on fine-grained feature interaction, i.e., the token-level intent features and slot features.
Previous works mainly focus on jointly modeling the relationship between the two subtasks with attention-based models, while ignoring the exploration of attention order.
In this paper, we propose to replace the conventional attention with our proposed Bilinear attention block and show that the introduced Higher-order Attention Network (HAN) brings improvement for the SLU task.
Importantly, we conduct wide analysis to explore the effectiveness brought from the higher-order attention.
 \end{abstract}

\section{Introduction}
Intent detection (ID) and Slot filling (SF) play important roles in SLU system. For instance, given an utterance "\textit{I want to listen to Hey Jude}", ID can be seen as a classification task to identity the user's intent is to listen to a song and SF can be treated as a sequence labeling task to produce a slot label sequence in BIO format \citep{ramshaw1999text,zhang2016joint} which demonstrate that \textit{Hey Jude} is the song’s title. Table \ref{tab:my-table} shows the expected output of SLU system for this instance.


Taking into account the relationship between these two tasks, joint modeling of them has gradually become the dominant method recently  \citep{goo2018slot,liu2019cm,niu2019novel,qin2019stack,qin2020co, huang2020FLSLU, Zhou2020PINAN, huang2021sentiment}.
The state-of-the-art methods \citep{li2018self,qin2019stack, qin2020co} adopt the attention mechanism \cite{vaswani2017attention} to trigger the mutual interaction between intent features and slot features.
Concretely, the attention mechanism learns a set of weights which reflect the importance of different words of an utterance via linearly fusing the given query and key via element-wise sum, and the weights are then applied to the value to derive a weighted sum which represents the enhanced intent or slot representation in a co-interactive way.

\begin{table}[t]
\centering
\scriptsize
\begin{tabular}{|c|c|c|c|c|c|c|c|}
\hline
\textbf{Utter.} & I & want&to&listen&to&Hey&Jude \\ \hline
\textbf{Slot}      & O&O&O&O&O&B-SONG&I-SONG \\ \hline
\textbf{Intent}    & \multicolumn{7}{c|}{PLAY\_SONG} \\ \hline
\end{tabular}
\caption{Example of SLU output for an utterance. Slot labels are in BIO format: B indicates the start of a slot span, I indicates the inside of a span while O denotes that the word does not belong to any slot.}
\label{tab:my-table}
\vspace{-3pt}
\end{table}
\begin{figure*}[t]
\parbox{.4\linewidth}{
    \centering
    \includegraphics[width=0.4\textwidth]{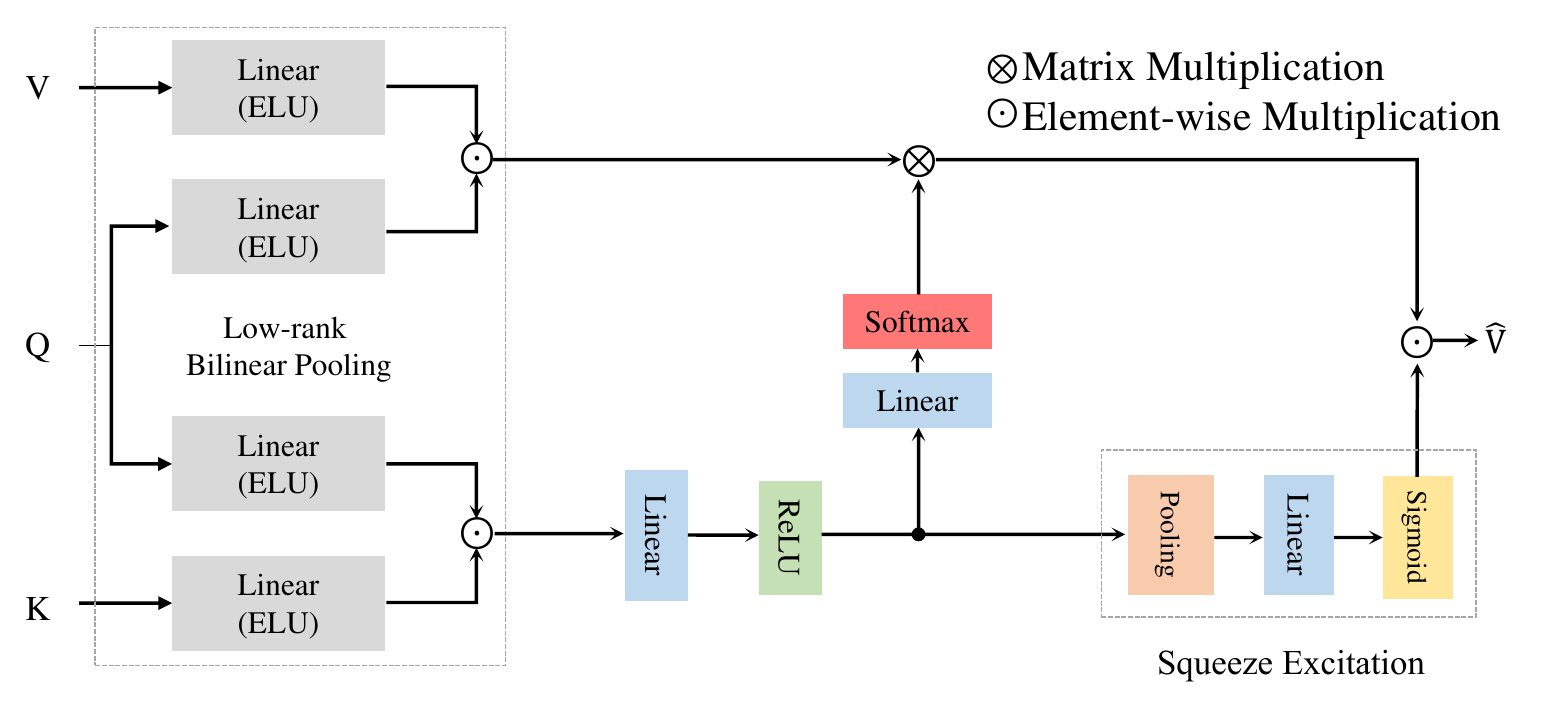}
    \caption{The proposed BiLinear attention block, which is based on the low-rank bilinear pooling \cite{kim2016hadamard}. }
	\label{fig:bilinear}
}
\hfill
\parbox{0.6\linewidth}{
    \centering
    \includegraphics[width=1\linewidth]{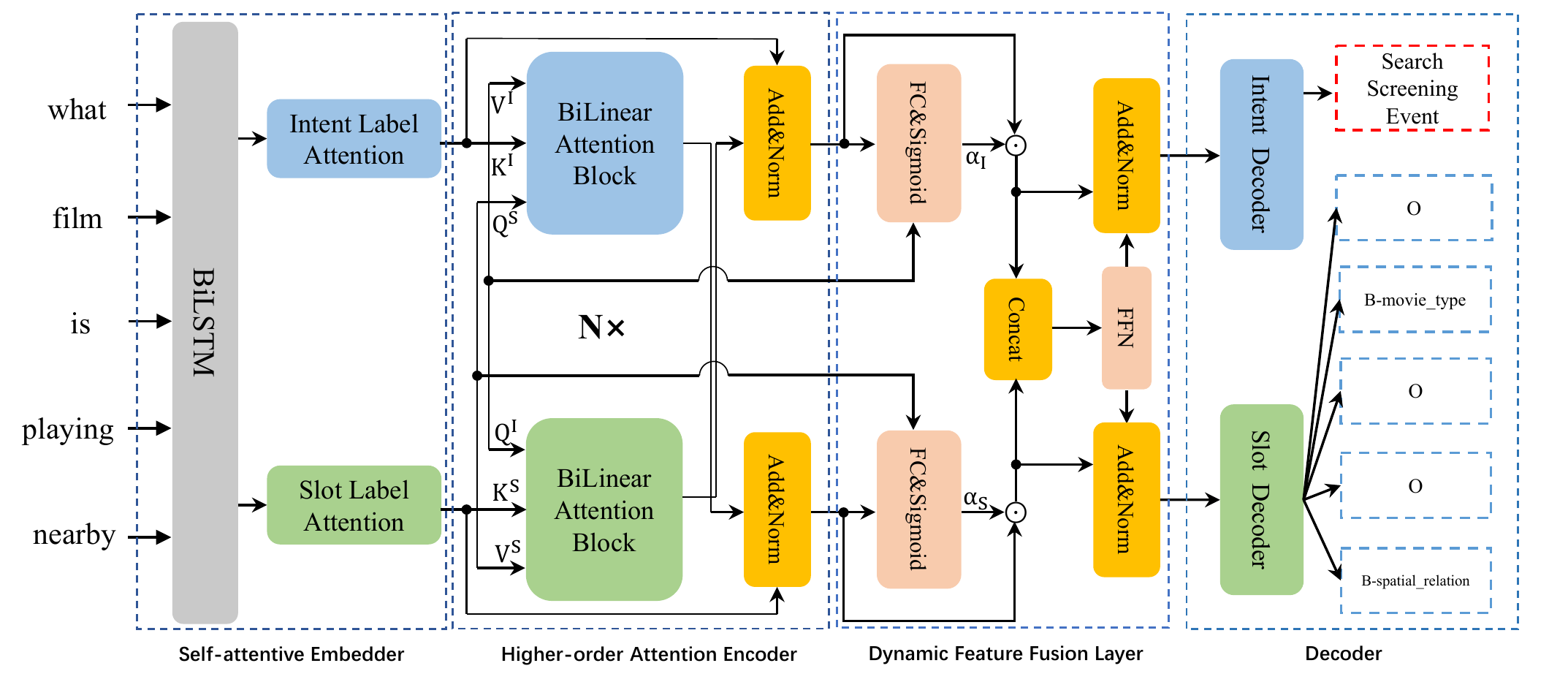}
    \vspace{-25pt}
    \caption{Architecture of the proposed HAN. 
    }
	\label{fig:model}
}
\end{figure*}

In this paper, we argue that the inherent design of conventional attention mechanism can only model the $1^{st}$ order feature interaction between query and key, which, however, is inefficient to model the relationship between ID and SF. 
Therefore, to obtain more representative intent and slot features, we propose to exploit higher-order interactions from $2^{nd}$ order feature interaction via bilinear pooling, which is an operation to calculate outer product between two feature vectors. Such
technique can enable the $2^{nd}$ order feature interaction by taking all pair-wise interactions between query and key into account and thus provide more discriminative representations. \citet{lin2015bilinear} first applied bilinear pooling to fuse visual features for fine-grained visual recognition. In order to mitigate the high computational complexity of bilinear pooling, \citet{kim2016hadamard} proposed low-rank bilinear pooling with linear mapping and Hadamard product.
Inspired by the successful application of bilinear pooling in the field of Computer Vision research, we proposed our BiLinear attention block as shown in Figure \ref{fig:bilinear}, which can build the
$2^{nd}$ order interactions between intent and slot features and get more discriminative intent and slot representations. Intuitively, a stack of the blocks is readily grouped to go beyond bilinear models and extract higher-order interactions. 
To this end, we also provide the view of how to integrate such blocks into HAN for building higher-order feature interactions.
The experiments show that our model achieves new state-of-the-art results on two benchmark datasets SNIPS \citep{Coucke2018Snips} and ATIS \citep{Hemphill1990ATIS}.


\begin{table*}[t]
\centering
\scriptsize
	\begin{tabular}{l c c c c c c}
		\toprule
        \multirow{2}{*}[-3pt]{\textbf{Model}} 
		&  \multicolumn{3}{c}{\textbf{SNIPS}} & \multicolumn{3}{c}{\textbf{ATIS}} \\ 
		\cmidrule(lr){2-4} \cmidrule(lr){5-7}  & Slot (\textit{F1})&Intent (\textit{Acc}) & Overall (\textit{Acc}) & Slot (\textit{F1})& Intent (\textit{Acc}) & Overall (\textit{Acc}) \\
		\midrule [\heavyrulewidth]
		Stack-Propagation \cite{qin2019stack} & 94.20 & 98.00 & 86.90 & 95.90 & 96.90 & 86.50 \\
		Co-Interactive \cite{qin2020co} & 95.35 & 98.71 & 89.12 & 95.47 & 97.65 & 86.69 \\ 
		Graph-LSTM \cite{zhang2020graph} & 95.30 & 98.29 & 89.71 & 95.91 & 97.20 & 87.57 \\
		\midrule [\heavyrulewidth]
		Baseline (BiLSTM+Decoder) & 94.19 & 97.79 & 85.86 & 95.32 & 95.63 & 84.99 \\
		\ + (Label attention+shallow concat)  & 94.39 & 98.03 & 87.89 & 95.55 & 97.52 & 85.89 \\
        \ + Conventional Attention ($1^{st}$ order based model) & 95.37 & 98.34 & 88.12 & 95.64 & 97.43 & 87.01\\
		\ + Bilinear attention block & 95.35 & 98.43 & 88.57 & 95.83 & 97.43 & 87.32 \\
		\ + Dynamic Feature Fusion & 95.57 & 98.57 & 89.43 & 95.88 & 97.56 & 87.57 \\
		\ + ELU & 96.01 & 98.69 & 90.43 & 95.95 & 97.89 & 88.12 \\ \midrule [\heavyrulewidth]
		HAN (Ours) & \textbf{96.18} & \textbf{99.12} & \textbf{91.80} & \textbf{96.12}& \textbf{98.04} & \textbf{88.47}\\
		HAN w/ BERT & 97.66 & 99.23 & 93.54 & 96.83& 98.54 & 89.31\\
		\midrule [\heavyrulewidth]
	\end{tabular}
	\vspace{-10pt}
    \caption{\label{experiment} Performance of different model on the SNIPS and ATIS datasets. Our HAN achieves the state-of-the-art performance on the two benchmark datasets.}
    \vspace{-10pt}
\end{table*}

\section{Approach}
We briefly formulate our HAN (Figure~\ref{fig:model}) which consists of four parts and introduce our BiLinear attention block in detail.

\subsection{BiLinear attention block}
In this section, we will describe the proposed Bilinear attention block as shown in Figure \ref{fig:bilinear} in detail.

Supposed we have a query $\mathbf{q}\in\mathbb{R}^{d}$, a set of keys ${\mathbf{K}=\{\mathbf{k}_i\}}_{i=1}^n$, and a set of values ${\mathbf{V}=\{\mathbf{v}_i\}}_{i=1}^n$, where $\mathbf{k}_i, \mathbf{v}_i \in \mathbb{R}^{d}$ denote the $i$-th key/value pair. Our block first performs low-rank bilinear pooling \cite{kim2016hadamard} to achieve a joint bilinear query-key representation $\mathbf{B}_i^k \in \mathbb{R}^d$ to model the $2^{nd}$ order feature interactions between query and key:
    $\mathbf{B}_i^k = {\rm ReLU}(\mathbf{W}_k\mathbf{k}_i)\odot{\rm ReLU}(\mathbf{W}_q^k\mathbf{q})$,
where $\mathbf{W}_k, \mathbf{W}_q^k\in\mathbb{R}^{d\times d}$ are weight matrices.

Next, depending on all bilinear query-key representations ${\{\mathbf{B}_i^k\}}_{i=1}^n$, two kinds of bilinear attention distributions are obtained to aggregate both contextual and channel-wise information within all values. Specifically, the contextual bilinear attention distribution is introduced by projecting each bilinear query-key representation into the corresponding attention weight via two embedding layers, followed with a softmax layer for normalization:
\begin{equation}
\footnotesize
\begin{aligned}
    \mathbf{B}_i^{'k} = {\rm ReLU}&(\mathbf{W}^k_B\mathbf{B}_i^{k}); \ b_i^s = \mathbf{W}_b\mathbf{B}_i^{'k}; \\ \mathbf{\beta^s} =& softmax(\mathbf{b^s}),
\end{aligned}
\end{equation}
where $\mathbf{W}^k_B\in\mathbb{R}^{d\times d}$ and $\mathbf{W}_b\in\mathbb{R}^{1\times d}$ are weight matrices, $\mathbf{B}_i^{'k}$ is the transformed bilinear query-key representation, and $b_i^s$ is the $i$-th element in $\mathbf{b^s}$. Here each element $\mathbf{\beta}^s_i$ in $\mathbf{\beta^s}$ denotes the normalized contextual attention weight for each key/value pair. Meanwhile, we perform
a squeeze-excitation operation over over all transformed
bilinear query-key representations $\{\mathbf{B}_i^{'k}\}_{i=1}^n$ for channel-wise attention measurement. Concretely, the operation of
squeeze aggregates all transformed bilinear query-key representations via average pooling, leading to a global channel descriptor
    $\overline{\mathbf{B}} = \frac{1}{n}\sum\limits_{i=1}^n{\mathbf{B}_i^{'k}}$.
After that, the followed excitation operation produces channel-wise attention distribution $\mathbf{\beta^c}$ by by leveraging the self-gating mechanism with a sigmoid over the $\overline{\mathbf{B}}$:
\begin{equation}
\footnotesize
    \mathbf{b^c} = \mathbf{W}_e\overline{\mathbf{B}}, \mathbf{\beta^c} = \sigma(\mathbf{b^c}),
\end{equation}
where $\mathbf{W}_e\in\mathbb{R}^{d\times d}$ is weight matrix.

Finally, our BiLinear attention block generates the attended value feature $\hat{\mathbf{v}_i}$ by accumulating the enhanced bilinear values with contextual and channel-wise bilinear attention:
\begin{equation}
\footnotesize
\begin{aligned}
    \hat{\mathbf{v}_i} &= \mathbf{\beta^c} \odot \sum\limits_{i=1}^n\mathbf{\beta}^s_i\mathbf{B}^v_i,\\
    \mathbf{B}^v_i &= {\rm ReLU}(\mathbf{W}_v\mathbf{v}_i)\odot{\rm ReLU}(\mathbf{W}_q^v\mathbf{q}_i),
\end{aligned}
\end{equation}
where $\mathbf{B}^v_i$ denotes the enhanced value of bilinear pooling on query $\mathbf{q}_i$ and each value $\mathbf{v}_i$, $\mathbf{W}_v\in\mathbb{R}^{d\times d}$ and $\mathbf{W}_q^v\in\mathbb{R}^{d\times d}$ are weight matrices. As such, BiLinear attention block produces more representative attended feature since higher-order feature interactions are exploited via bilinear pooling. We iterate the above process $n$ times with ${\mathbf{Q}=\{\mathbf{q}_i\}}_{i=1}^n$, and get a set of values ${\mathbf{\hat{V}}=\{\hat{\mathbf{v}_i}\}}_{i=1}^n$:
\begin{equation}
\footnotesize
    \mathbf{\hat{V}} = F_{BiLinear}(\mathbf{K, V, Q}),
\end{equation}
where $\mathbf{\hat{V}}\in\mathbb{R}^{n\times d}$ is the enhanced features with higher-order attention feature interactions. By equipping the block with \textit{Exponential Linear Unit (ELU)}\cite{barron2017continuously}, it can model infinity order feature interactions, which can be proved via Taylor expansion of each element in bilinear vector after exponential transformation.
Specifically, for two vectors X and Y, their exponential bilinear pooling can be estimated using the Taylor expansion:
\begin{equation*}
\label{equ:taylor}
\tiny
\begin{aligned}
&\exp \left(\mathrm{W}_{X} \mathrm{X}\right) \odot \exp \left(\mathrm{W}_{Y} \mathrm{Y}\right) \\
&=\left[\exp \left(\mathrm{W}_{X}^{1} \mathrm{X}\right) \odot \exp \left(\mathrm{W}_{Y}^{1} \mathrm{Y}\right), \ldots, \exp \left(\mathrm{W}_{X}^{D} \mathrm{X}\right) \odot \exp \left(\mathrm{W}_{Y}^{D} \mathrm{Y}\right)\right] \\
&=\left[\exp \left(\mathrm{W}_{X}^{1} \mathrm{X}+\mathrm{W}_{Y}^{1} \mathrm{Y}\right), \ldots, \exp \left(\mathrm{W}_{X}^{D} \mathrm{X}+\mathrm{W}_{Y}^{D} \mathrm{Y}\right)\right] \\
&=\left[\sum_{p=0}^{\infty} \gamma_{p}^{1}\left(\mathrm{~W}_{X}^{1} \mathrm{X}+\mathrm{W}_{Y}^{1} \mathrm{Y}\right)^{p}, \ldots, \sum_{p=0}^{\infty} \gamma_{p}^{D}\left(\mathrm{~W}_{X}^{D} \mathrm{X}+\mathrm{W}_{Y}^{D} \mathrm{Y}\right)^{p}\right],
\end{aligned}
\end{equation*}
where ${\rm{W}}_X$ and ${\rm{W}}_Y$ are embedding matrices, $D$ denotes the dimension of bilinear vector, ${\rm{W}}^i_X$/${\rm{W}}^i_Y$ is the $i$-th row in ${\rm{W}}_X$/${\rm{W}}_Y$.  

\subsection{Overview of the HAN}

\smallskip\noindent\textbf{Self-attentive Embedder} \
Inspired by \citet{qin2019stack}, we employ the Self-attentive Embedder to obtain the utterance embeddings.~\nocite{audio2021MRC}
It first uses a shared BiLSTM to embed the input sequence, acquiring $\mathbf{H} = (\mathbf{h}_1, \mathbf{h}_2,..., \mathbf{h}_n)$.
Then, it perform the label attention \cite{cui2019hierarchically} over intent and slot label to get the explicit intent and slot representation $\mathbf{H_I}\in \mathbb{R}^{n\times d}$, $\mathbf{H_S}\in \mathbb{R}^{n\times d}$ ($d=128$), which capture the intent and slot semantic information, respectively.



\smallskip\noindent\textbf{Higher-order Attention Encoder} \
$\mathbf{H_I}$ and $\mathbf{H_S}$ are further fed into the our Higher-order Attention Encoder to strengthen both the intent and slot features via capturing higher-order feature interactions between them.
Formally, the encoder is composed of a stack of $N=2$ identical sublayers. Each sublayer is our Bilinear attention block, followed by layer normalization. Same with \citet{vaswani2017attention}, we first map the matrix $\mathbf{H_I}$ and $\mathbf{H_S}$ to queries$\mathbf{(Q_I^{(1)}, Q_S^{(1)})}$, keys$\mathbf{(K_I^{(1)}, K_S^{(1)})}$ and values$\mathbf{(V_I^{(1)}, V_S^{(1)})}$ matrices by using different linear projections. Then we take $\mathbf{Q_I^{(1)}}$, $\mathbf{K_S^{(1)}}$ and $\mathbf{V_S^{(1)}}$ as queries, keys and values, respectively, acquiring the enhanced values:
\begin{equation}
\footnotesize
\begin{aligned}
\label{equ:attention}
    \hat{\mathbf{V_I^{(1)}}} =& F_{BiLinear}(\mathbf{K_S^{(1)}}, \mathbf{V_S^{(1)}}, \mathbf{Q_I^{(1)}}), \\
    \mathbf{H_I^{(1)}} =& {\rm LN}(\mathbf{H_I}+\hat{\mathbf{V_I^{(1)}}}),
\end{aligned}
\end{equation}
where ${\rm LN}$ represents the layer normalization \cite{ba2016layer}. Similarly, we take $\mathbf{Q_S^{(1)}}$ as queries, $\mathbf{K_I^{(1)}}$ as keys and $\mathbf{V_I^{(1)}}$ as values to obtain $\mathbf{H_S^{(1)}}$.

After repeating $N$ times, we can obtain the enhanced intent features $\mathbf{H_I^{(N)}}\in\mathbb{R}^{n\times d}$ and slot features $\mathbf{H_S^{(N)}}\in\mathbb{R}^{n\times d}$, which are endowed with the higher-order feature interactions in between.

\smallskip\noindent\textbf{Dynamic Feature Fusion Layer} \
We first compute two weight matrices $\mathbf{\alpha_I}$ and $\mathbf{\alpha_S}$ which reflect the relevance between the output and the input query of the last sublayer in the Higher-order Attention Encoder, and thus obtain the fused features $\mathbf{H_{IS}}$, which can be defined as follows:
\begin{equation}
\footnotesize
\begin{aligned}
    \mathbf{\alpha_I} =& \sigma(W_I\mathbf{[Q_I^{(N)}, H_I^{(N)}]}+ b_I), \\ 
    \mathbf{\alpha_S} =& \sigma(W_S\mathbf{[Q_S^{(N)}, H_S^{(N)}]} + b_S), \\
    \mathbf{H_{IS}} =& \mathbf{\alpha_I\odot H_I^{(N)} \oplus \alpha_S\odot H_S^{(N)}},
\end{aligned}
\end{equation}
where $[\cdot,\cdot]$ indicates concatenation, $\sigma$ is the sigmoid activation; $\odot$ denotes element-wise multiplication; $W_I$ and $W_S$ are both $2d\times d$ embedding matrices, $b_I$ and $b_S$ are biases.
Then, we adopt the feed-forward network (FFN) \cite{zhang2016joint}, to acquire the updated intent features $\mathbf{\hat{H_I}^{(N)}}\in \mathbb{R}^{n\times d}$ and slot features $\mathbf{\hat{H_S}^{(N)}}\in \mathbb{R}^{n\times d}$, i.e., $\mathbf{\hat{H_I}^{(N)}} =  {\rm LN}(\mathbf{\mathbf{FFN}(\mathbf{H_{IS}})+H_I^{(N)}})$ and $\mathbf{\hat{H_S}^{(N)}} =  {\rm LN}(\mathbf{\mathbf{FFN}(\mathbf{H_{IS}})+H_S^{(N)}})$.



\smallskip\noindent\textbf{SLU Decoder} \
For the intent detection, we follow \citet{kim2014convolutional} to employ the \textit{maxpooling} on $\mathbf{\hat{H_I}^{(N)}}$ to obtain $\boldsymbol c$, which is used to predict the intent label: 
    $\mathbf{o^I} \sim \mathbf{\hat{y}^I} = \rm softmax\left(\mathbf{W}^{\mathbf{I}} \boldsymbol c+\mathbf{b_I}\right)$,

For the slot filling, we apply a standard CRF layer \cite{niu2019novel} to model the dependency between labels, and then predict the label sequence $P(\hat{y}|\mathbf{O_S}) = \frac{\sum\limits_{i=1}{\rm exp} f(y_{i-1},y_i,\mathbf{O_S})}{\sum\limits_{y^{\prime}}\sum\limits_{i=1}{\rm exp}f(y_{i-1}^{\prime},y_i^{\prime},\mathbf{O_S})}$, where $f(y_{i-1}^{\prime},y_i^{\prime},\mathbf{O_S})$ computes the transition score from $y_{i-1}$ to $y_{i}$ and $\mathbf{O_S} = \mathbf{W^S}\mathbf{\hat{H_S}^{(N)}}+\mathbf{b_S}$.


\section{Experiments} 


\smallskip\noindent\textbf{Main Results} \
Table~\ref{experiment} shows the results of our approach on the SNIPS and ATIS, our HAN outperforms all baselines and achieves the state-of-the-art performance. Besides, fine-tuned with the strong pre-trained language model (BERT)~\cite{Devlin2019BERT}~\nocite{hou2021DynaBERT, huang2021GhostBERT}, HAN has been further improved. For the ablation study, we can see that as adding each key component of the model gradually, the performance gradually becomes better, and it gets improvement when equipping with ELU (4.57\% and 3.13\% improvement compared to baseline in overall accuracy on the SNIPS and ATIS dataset, respectively).
The "shallow concat" means directly concatenate the two features without dynamic feature fusion mentioned above. 






\smallskip\noindent\textbf{Robustness towards Learning Rate} \
\label{sec:lr}
\begin{figure}
    \centering
    \includegraphics[width=0.9\linewidth]{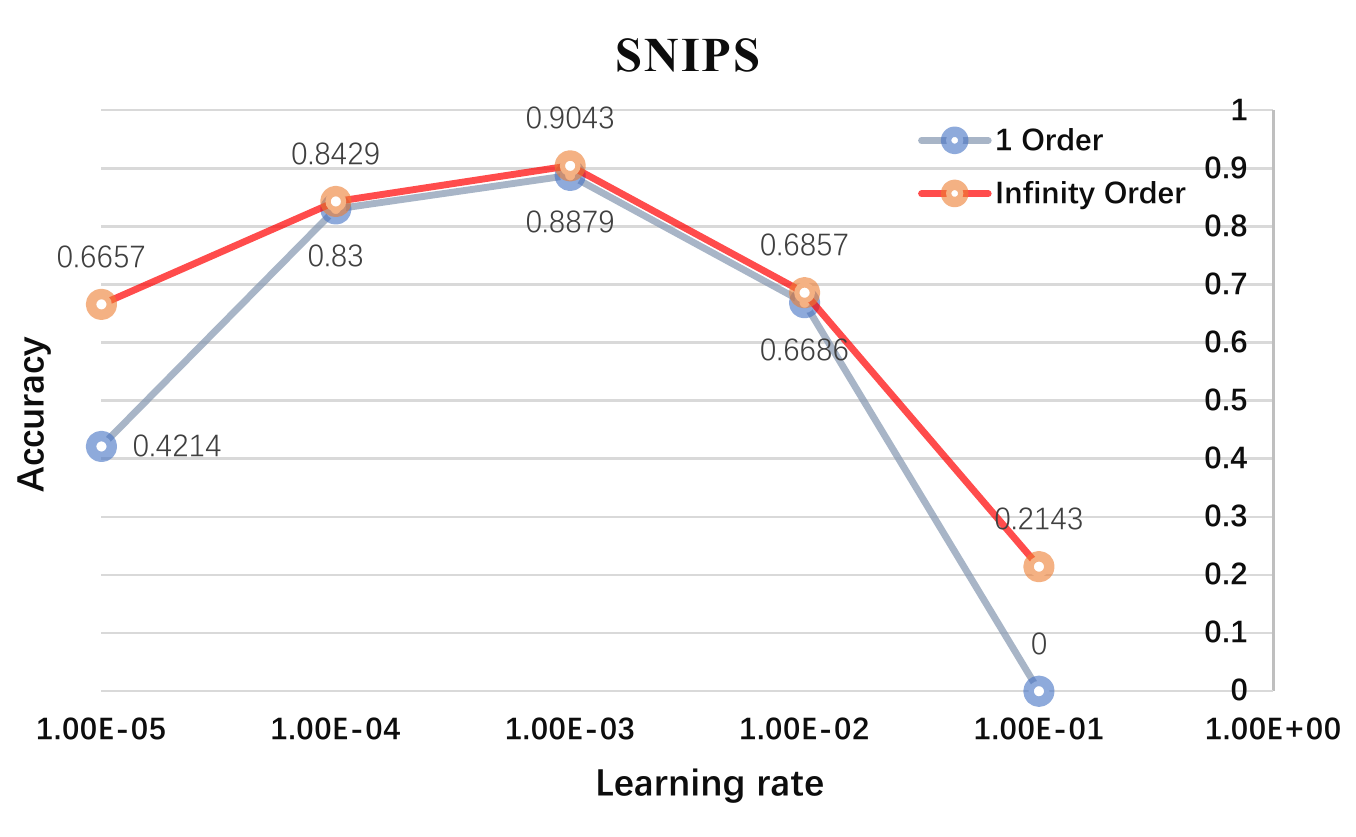}
    \vspace{-7pt}
    \caption{Performance of the SLU model w/ ELU (HAN) and w/o ELU ($1^{st}$ order) under different learning rate. The $1^{st}$ order based model is acquired by replacing our higher-order attention with the conventional attention model \cite{vaswani2017attention}.}
	\label{fig:lr_snips}
	\vspace{-10pt}
\end{figure}
From Figure~\ref{fig:lr_snips}, with the overall Accuracy as metric, we show that under same experimental settings, infinity order SLU model performs better than the first order model under different learning rate.
Besides, we find that under reasonable and task-specific range of the learning rate, i.e., 1e-4 to 1e-2, the higher-order based model performs slightly and consistently better than the $1^{st}$ order based model.
While when the learning rate is out of this range, i.e., bigger than 1e-2 or smaller than 1e-4, the performance of the $1^{st}$ order based model occurs to crack down quickly and it even drops to 0 when the learning rate is 0.1.
Oppositely, the higher-order based model, though also drops down when the learning reaches is set out of reasonable range, can still keep considerable performance compared to the first order base model, and thus shows its robustness to the extreme case towards the optimized learning rate.
A similar phenomenon also can be found on the ATIS dataset (please see our Appendix).

\smallskip\noindent\textbf{Generalization Analysis} \
\label{generalization}
We attempt to incorporate the Higher-order Attention Encoder (marked as HAE) into several existing baselines. Table~\ref{tab:generalization} shows that baselines with infinity order attention, i.e., HAE, performs better than it with $1^{st}$ order attention model in most of the case.
This further verifies the generalization of the effectiveness of the higher-order attention on the SLU task.

\begin{table}[t]
\scriptsize
\centering
\setlength{\tabcolsep}{3pt}
\begin{tabular}{l c c c c c c c}
	\toprule
	\multirow{2}{*}[-3pt]{Model} & \multirow{2}{*}[-3pt]{HAE}
	&  \multicolumn{3}{c}{\textbf{SNIPS}}  &  \multicolumn{3}{c}{\textbf{ATIS}}\\
	\cmidrule(lr){3-5} \cmidrule(lr){6-8}
	& & Slot &Intent & Overall & Slot &Intent & Overall\\
	\midrule [\heavyrulewidth]
	\multirow{2}{*}{\begin{tabular}[c]{@{}l@{}} Stack-Propagation \\ \cite{qin2019stack}  \end{tabular}} & $\times$ & 94.20 & 98.00 & 86.90 & 95.90 & 96.90 & 86.50 \\
	& $\checkmark$ & \textbf{94.70} & \textbf{98.20} & \textbf{87.20} & \textbf{96.20} & \textbf{97.50} & \textbf{87.40} \\
	\midrule
	\multirow{2}{*}{\begin{tabular}[c]{@{}l@{}} Co-Interactive \\ \cite{qin2020co} \end{tabular}} & $\times$ & 95.35 & \textbf{98.71} & 89.12 & 95.47 & 97.65 & 86.69  \\
	& $\checkmark$ & \textbf{95.49}  & 98.57 & \textbf{89.86} & \textbf{95.98} & \textbf{97.87} & \textbf{87.35} \\
	\midrule
    \multirow{2}{*}{HAN (Ours)} & $\times$ & 95.43 & 98.57 & 89.29 & 95.87 & 97.42 & 87.12 \\
	& $\checkmark$ & \textbf{96.18} & \textbf{99.12} & \textbf{91.80} & \textbf{96.12} & \textbf{98.04} & \textbf{88.47} \\
	\bottomrule
\end{tabular}
\caption{\label{tab:generalization}Impact of higher-order attention on different baselines, i.e., Stack Propagation, Co-Interactive, and our HAN.}
\vspace{-10pt}
\end{table}

\begin{figure}[t]
	\noindent\begin{minipage}{.99\linewidth}
		\centering
		\includegraphics[width=0.99\linewidth]{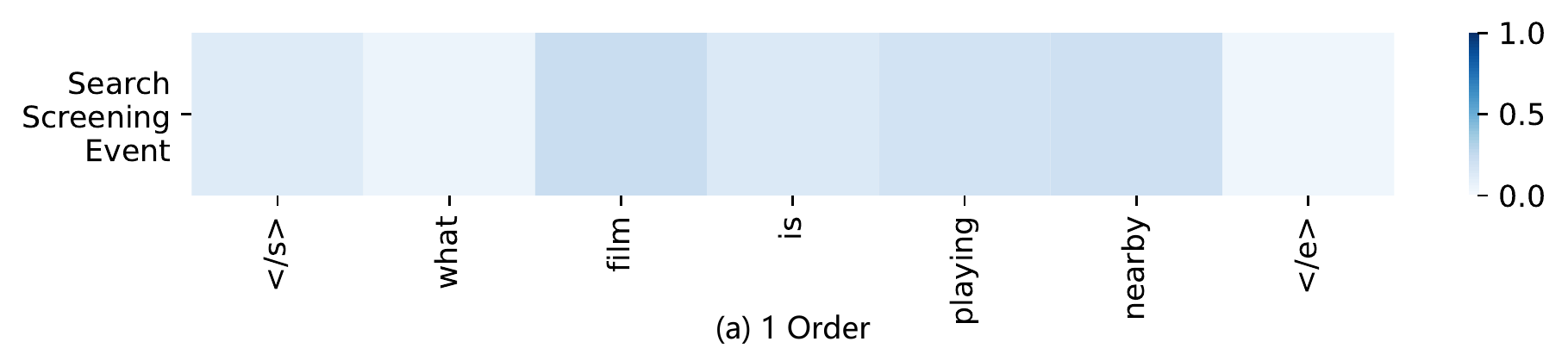}
	\end{minipage}
	\noindent\begin{minipage}{.99\linewidth}
		\centering
		\includegraphics[width=0.99\linewidth]{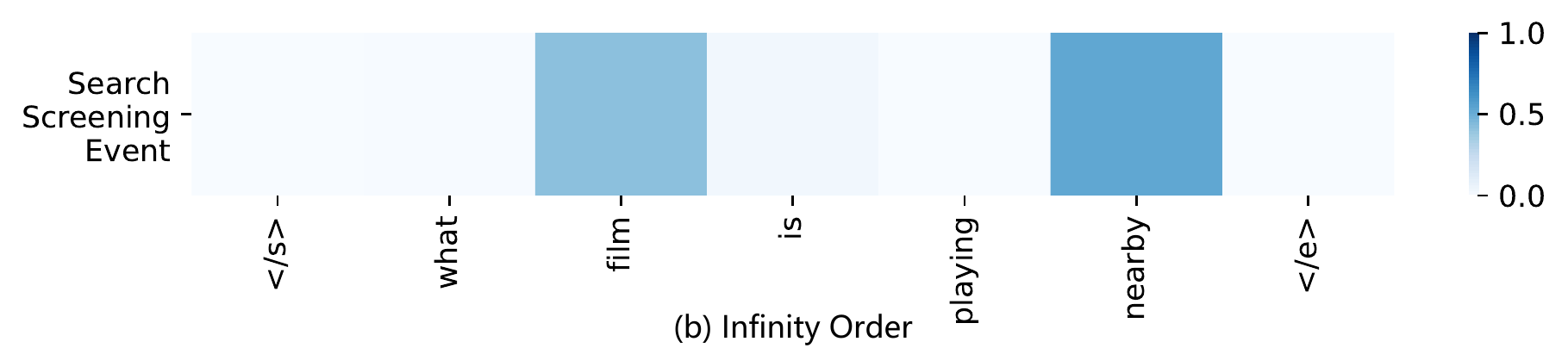}
		\captionof{figure}{Visualization of our proposed methods. The utterance is acquired from the SNIPS dataset. 
		}
	\label{fig:visualization}
	\end{minipage}
	\vspace{-10pt}
\end{figure}


\smallskip\noindent\textbf{Visualization of the Infinity Order Attention} \
To explore the model promotion brought from the proposed HAN, we turn to visualize the attention pattern over Q and K in Eq.~(\ref{equ:attention}).
As can be seen in Figure~\ref{fig:visualization}, where we use the utterance "\textit{What film is playing nearby}" from the SNIPS dataset,
the proposed higher-order attention model shows clearer attention capture ability compared to the original $1^{st}$ attention mechanism, it can better focus on the keyword \textit{film} and \textit{nearby}.

\section{Conclusion}

In this paper, we propose a novel Bilinear attention block which can build the $2^{nd}$ order interactions between intent and slot features and get more discriminative intent and slot representations. The higher and even infinity order feature interactions can be readily modeled via stacking multiple BiLinear attention
blocks and equipping the block with ELU activation. Moreover, we introduced the HAN, and conducted numerous experiments and analysis on SNIPS and ATIS datasets to demonstrate the effectiveness of our method.




\appendix
\section{Related Work}

\subsection{Spoken Language Understanding}
Spoken Language Understanding is a well known task in dialogue system, and it typically contains intent detection and slot filling tasks.
The special relation of the two tasks requires them to have enough correlation and interaction, making it possible to explore the promotion brought from higher order attention.


\subsection{Bilinear Pooling}
Bilinear pooling was first proposed in~\cite{lin2015bilinear} to fuse the features for fine-grained visual recognition, it can provide $2^{nd}$ order interaction on feature vectors.
Later for the SLU task,
~\citet{Teng2020ICASSP} proposed to use Bilinear pooling to fuse the word information and character information, showing that such method can provide more discriminative representations than simple pooling, i.e., $1^{st}$ order, for the spoken language understanding task.

\section{Experimental Details}
\label{sec:appendix_exp}
\subsection{Experimental Settings}
We adopt the RAdam \citep{liu2019variance} optimizer for optimizing the parameters, with a mini-batch size of 32 and initial learning rate of 0.001. We use 300d GloVe pre-trained vector~\cite{pennington2014glove} as the initialization embedding. The hidden dimensionality is set as 128.
Two evaluation metrics are used in the SLU task. The performance of intent detection is measured by accuracy, while slot filling is evaluated with the F1 score, and the sentence-level semantic frame parsing using overall accuracy.

\section{More Experiment Results}

\subsection{Learning Rate on ATIS}
\label{sec:appendix_lr}
\begin{figure}
    \centering
    \includegraphics[width=0.99\linewidth]{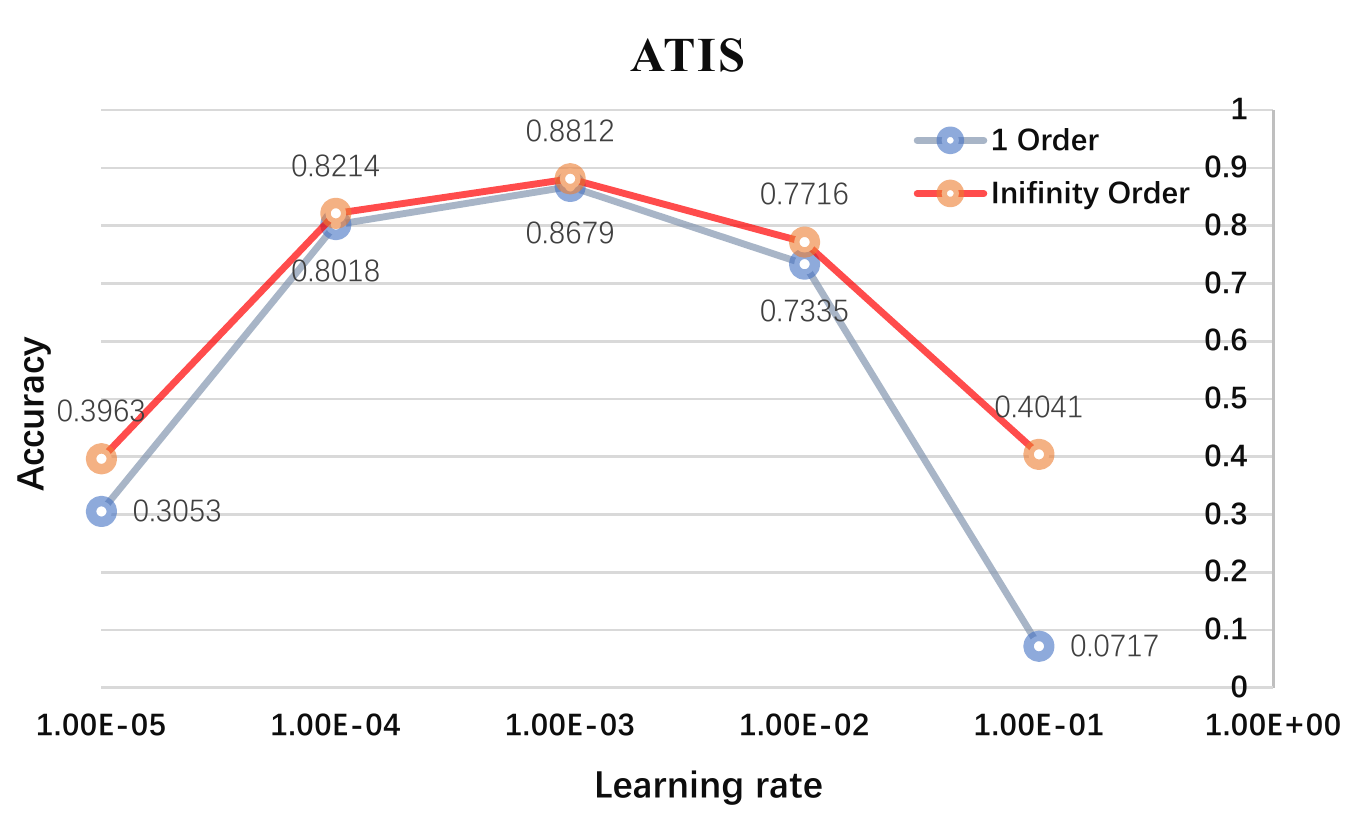}
    \caption{Performance of the model w/ELU and wo/ELU under different learning rate on the ATIS dataset.}
	\label{fig:lr_atis}
\end{figure}

Besides the performance of the higher order model and the first order model in the SNIPS dataset w.r.t, learning rate in Section~\ref{sec:lr}, we also show the results in the ATIS dataset.
From Figure~\ref{fig:lr_atis}, we can see that the higher order attention based model performs better than the first order based model consistently, and it shows a clear robustness towards the learning rate for its adaptive capacity in the hyper-parameter.

\subsection{Discussion on the layer number of Higher-order Attention Encoder}

\begin{table}[t]
\small
\begin{center}
\setlength{\tabcolsep}{3pt}
	\begin{tabular}{c c c c c c c c}
		\toprule
		\multirow{2}{*}[-3pt]{L} & \multirow{2}{*}[-3pt]{ELU} 
		&  \multicolumn{3}{c}{\textbf{SNIPS}} & \multicolumn{3}{c}{\textbf{ATIS}} \\ 
		\cmidrule(lr){3-5}  \cmidrule(lr){6-8}
		& & Slot &Intent & Overall & Slot & Intent& Overall \\
		\midrule [\heavyrulewidth]
        \multirow{2}{*}{1} & $\times$ & 95.57 &98.57& 89.43 &95.88 &97.56 &	87.57 \\
        & \checkmark & 96.01 &98.69 &90.43 &95.95 &	97.89&	88.12\\ \midrule   
        
        \multirow{2}{*}{2} & $\times$ & 95.86&	98.57&89.71&95.92&97.76&	87.79\\
        & \checkmark & 96.18 &99.12 &91.80 &96.12 &	98.04 &	88.47 \\ \midrule
        
        \multirow{2}{*}{3} & $\times$ &95.65&	98.29&89.57 &95.75 &97.54&	87.12\\
        & \checkmark &95.70&	98.57&89.86 &95.75 &97.42&87.23 \\ \midrule
        
        \multirow{2}{*}{4} & $\times$ & 95.51&	98.00&89.00 &95.75 &	97.31 &	87.01 \\
        & \checkmark & 95.65&	98.29&89.11&95.77&	97.42 &	87.01 \\ \midrule
        
        \multirow{2}{*}{5} & $\times$ &95.12&97.86&	88.86 &95.58 &	97.54 &	86.67 \\ 
        & \checkmark & 95.32&98.14&	88.86 &95.83 &	97.31 &	86.90\\ 
		\bottomrule
	\end{tabular}
    \caption{\label{order} Comparison of different order model.}
\end{center}
\end{table}
From Table~\ref{order}, we can see that when the number of sublayers in Higher-order Attention Encoder is 2, performance of HAN on both validation dataset gets the best. So we set the number of sublayers $N=2$. We speculate that the increased parameters by stacking more blocks might result in overfitting, which somewhat hinders the exploitation of higher order interaction in this way.

\bibliography{emnlp2021}
\bibliographystyle{emnlp2021}

\end{document}